\documentclass[manuscript,screen,authorversion]{acmart}
\usepackage[linesnumbered,ruled,vlined]{algorithm2e} 
\usepackage{subfigure}
\usepackage{amsmath}
\DeclareMathOperator{\E}{\mathbb{E}}

\newcommand{\mcU}{\mathcal{U}}
\newcommand{\mcI}{\mathcal{I}}
\newcommand{\mcS}{\mathcal{S}}
\newcommand{\YT}{Y_{ui}^\textnormal{T}}
\newcommand{\YC}{Y_{ui}^\textnormal{C}}

\AtBeginDocument{%
  \providecommand\BibTeX{{%
    \normalfont B\kern-0.5em{\scshape i\kern-0.25em b}\kern-0.8em\TeX}}}

\setcopyright{acmcopyright}
\copyrightyear{2021}
\acmYear{2021}
\setcopyright{acmlicensed}\acmConference[RecSys '21]{Fifteenth ACM Conference on 
	Recommender Systems}{September 27-October 1, 2021}{Amsterdam, Netherlands}
\acmBooktitle{Fifteenth ACM Conference on Recommender Systems (RecSys '21), 
	September 27-October 1, 2021, Amsterdam, Netherlands}
\acmPrice{15.00}
\acmDOI{10.1145/3460231.3474235}
\acmISBN{978-1-4503-8458-2/21/09}

\begin{document}

\title{Online Evaluation Methods for the Causal Effect of Recommendations}

\author{Masahiro Sato}
\email{masatoh7373@gmail.com}
 \orcid{0000-0003-0000-6341}
 \affiliation{%
   \institution{Independent Researcher}
   \city{Kaisei}
   \state{Kanagawa}
   \country{Japan}
 }
\authornote{This is the author's version of the work. It is posted here for your personal use. Not for redistribution. The definitive version will be published in RecSys '21, https://doi.org/10.1145/3460231.3474235.}

\begin{abstract}
  Evaluating the causal effect of recommendations is an important objective because the causal effect on user interactions can directly leads to an increase in sales and user engagement.
  To select an optimal recommendation model, it is common to conduct A/B testing to compare model performance.
  However, A/B testing of causal effects requires a large number of users, making such experiments costly and risky.
  We therefore propose the first interleaving methods that can efficiently compare recommendation models in terms of causal effects.
  In contrast to conventional interleaving methods, we measure the outcomes of both items on an interleaved list and items not on the interleaved list, since the causal effect is the difference between outcomes with and without recommendations.
  To ensure that the evaluations are unbiased, we either select items with equal probability or weight the outcomes using inverse propensity scores.
  We then verify the unbiasedness and efficiency of online evaluation methods through simulated online experiments.
  The results indicate that our proposed methods are unbiased and that they have superior efficiency to A/B testing.
\end{abstract}

\ccsdesc[500]{Information systems~Recommender systems}
\ccsdesc[500]{Information systems~Retrieval effectiveness}

\keywords{causal inference, interleaving, A/B testing, treatment effect}

\maketitle

\section{Introduction}
A recommendation is a treatment that can affect user behavior.
An increase in user actions, such as purchases or views, by the recommendation is the treatment effect (also called the causal effect).
Because this leads to improved sales or user engagement, the causal effect of recommendations is important for businesses.
While most recommendation methods aim for accurate predictions of user behaviors, there may be a discrepancy between the accuracy and the causal effect of recommendations~\cite{Sato19}.
Several recent works have thus proposed recommendation methods to rank items by the causal effect of recommendations~\cite{Bodapati08,Sato16,Sato19,Sato20,Sato21}.

Online experiments are commonly conducted to compare model performance and select the best recommendation model.
However, evaluating the causal effect is not straightforward; we cannot naively compare the outcomes of recommended items because the causal effect is the difference between the \textit{potential outcomes} with and without the treatment~\cite{Rubin74,Imbens15}.
A/B testing that compares the total user actions on all items, not only recommended items, can reveal the difference in the average causal effect (see Section \ref{sec:AB_testing}).
Nevertheless, it suffers from large fluctuations due to the variability in natural user behaviors for non-recommended items: some users tend to purchase more items than others.
A large number of users is required to compensate for such fluctuations, making online experiments costly and risky.

In this paper, we propose efficient online evaluation methods for the causal effect of recommendations based on interleaving.
Interleaving generates a list from the lists ranked by the two models to be compared~\cite{Chapelle12}.
Whereas previous interleaving methods only measure the outcomes of items in the intersection of the original and interleaved lists, our proposed methods also measure the outcomes of items in the original lists but not in the interleaved list.
We propose an interleaving method that selects items with equal probability for unbiased evaluation.
With unequal selection probabilities, the evaluation might be biased due to confounding~\cite{Hernan20} between recommendation and potential outcomes, leading to inaccurate judgments of the recommendation models.
We remove the possible bias by properly weighting the outcomes based on the inverse propensity score (IPS) method used in causal inference~\cite{Rosenbaum83,Lunceford04}.
This enables the use of a more general interleaving framework that only requires non-zero probabilities to be selected for any item in the original lists.
As an instance of the framework, we propose a causal balanced interleaving method that balances the number of items chosen from the two compared lists.
To verify the unbiasedness and efficiency of the proposed interleaving methods, we simulate online experiments to compare ranking models.

The contributions of this paper are summarized as follows.
\begin{itemize}
	\item We propose the first interleaving methods to compare recommendation models in terms of their causal effect.
	\item We verify the unbiasedness and efficiency of the proposed methods through simulated online experiments.
\end{itemize}

\section{Related Work}
\subsection{Interleaving Methods}
Interleaving is an online evaluation method for comparing two ranking models by observing user interactions with an interleaved list that is generated from lists ranked by the two models to be compared~\cite{Chapelle12}.
Several interleaving methods have been proposed for evaluating information retrieval systems.
Balanced interleaving~\cite{Joachims02,Joachims03} generates an interleaved list from two rankings to be compared such that the highest ranks in the interleaved list $k_A$ and $k_B$ from the two rankings $A$ and $B$, respectively, are the same or different by at most one.
Team draft interleaving~\cite{Radlinski08} alternatively selects items from compared rankings, analogously to selecting teams for a friendly team-sports match.
Probabilistic interleaving~\cite{Hofmann11} selects items according to probabilities that depend on the item ranks.
Optimized interleaving~\cite{Radlinski13} makes the properties required for interleaving in information retrieval explicit and then generates interleaved lists by solving an optimization problem to fulfill those properties.
Interleaving methods have been extended to multileaving that compare multiple rankings simultaneously~\cite{Schuth14,Schuth15}.
Multileaving has been also applied to the evaluation of a news recommender system~\cite{Iizuka19}.
The objective of previous interleaving methods is to evaluate how accurately the rankings reflect queries or user preferences, whereas our goal is to evaluate rankings in terms of the causal effect.
To the best of our knowledge, at present there are no interleaving methods for causal effects.

\subsection{Recommendation Methods for the Causal Effect}
Recommendations can affect users' opinions~\cite{Cosley03} and induce users' actions~\cite{Dias08,Jannach19}.
However, users' actions on recommended items could have occurred even without the recommendations~\cite{Sharma15}.
Building recommendation models that target the causal effect is challenging because the ground truth data of causal effects are not observable~\cite{Holland86}.
One approach is to train prediction models for both recommended and non-recommended outcomes and then to rank the items based on the difference between the two predictions~\cite{Bodapati08,Sato16}.
Another approach is to optimize models directly for the causal effect.
ULRMF and ULBPR~\cite{Sato19} are respectively pointwise and pairwise optimization methods that use label transformations and training data samplings designed for causal effect optimization.
DLCE~\cite{Sato20} is an unbiased learning method for the causal effect that uses an IPS-based unbiased learning objective.
There are also neighborhood methods for causal effects~\cite{Sato21} that are based on a matching estimator in causal inference.
These prior works on causal effects evaluated methods offline and did not discuss protocols for online evaluation.
In this study, we develop online evaluation methods and compare some of the aforementioned recommendation methods in simulated online experiments.

Another line of works in the area of causal recommendation aims for debiasing~\cite{Chen20}.
Several methods have been proposed to learn users' true preferences from biased (missing-not-at-random) feedback data~\cite{Schnabel16,Saito20a,Wang20,Bonner18}.\footnote{Note that CausE  proposed by Bonner and Vasile~\cite{Bonner18} can be used for the causal effect ranking~\cite{Sato19}, although their original work tackles unbiased prediction of $\YT$ and only refers to the prediction of $\tau_{ui}$. Wang et al.~\cite{Wang20} suggested that they want to recommend items that have low probability of exposure and that would be rated high if exposed. Their approach might be regarded as indirectly targeting the causal effect of recommendations, assuming that recommendations increase exposures. To take this approach, it might be also important to model the influence of recommendations on exposures~\cite{Sato20b}.}
These methods can be regarded as predicting interactions with recommendations (i.e., $\YT$, defined in the next section).
Hence, we can evaluate them using previous interleaving methods.

\section{Evaluation Methods for the Causal Effect of Recommendations}
\subsection{Causal Effect of Recommendations}
In this subsection, we define the causal effect of recommendations.
Let $\mcU$ and $\mcI$ be sets of users and items, respectively.
Let $Y_{ui} \in \{0, 1\}$ denote the interaction (e.g., purchase or view) of user $u \in \mcU$ with item $i \in \mcI$.
User interactions may differ depending on whether the item is recommended or not.
We denote the binary indicator for the recommendation (also called the treatment assignment) by $Z_{ui} \in \{0, 1\}$.
Let $\YT$ and $\YC \in \{0, 1\}$ be hypothetical user interactions (also called potential outcomes \cite{Rubin74}) when item $i$ is recommended to $u$ ($Z_{ui} = 1$) and when it is not recommended ($Z_{ui} = 0$), respectively.
The causal effect $\tau_{ui}$ of recommending item $i$ to user $u$ is defined as the difference between the two potential outcomes: $\tau_{ui} = \YT - \YC,$
that takes ternary values, $\tau_{ui} \in \{-1, 0, 1\}$.
Using potential outcomes, the observed interaction can be expressed as
\begin{equation}
	Y_{ui} = Z_{ui}\YT + (1-Z_{ui})\YC.
\end{equation}
$Y_{ui} = \YT$ if $i$ is recommended ($Z_{ui} = 1$) and $Y_{ui} = \YC$ if it is not recommended ($Z_{ui} = 0$).
Note that $\YT$ or $\YC$ cannot both be observed at a specific time; hence, $\tau_{ui}$ is not directly observable.

The recommendation model $A$ generates a recommendation list $L_u^A$ for each user.
The average causal effect of model $A$ is then defined as
\begin{equation}
	\tau_{A} = \E [\tau_{ui}| i \in L_u^A, u \in \mcU].
\end{equation}
In this work, we evaluate models using the above metric.\footnote{This metric is identical to the causal precision@$n$ in \cite{Sato20}.}
That is, when comparing two models, we regard $A$ to superior to $B$ when $\tau_{A} > \tau_{B}$.

\subsection{A/B testing for the Causal Effect}
\label{sec:AB_testing}
For A/B testing, we randomly select non-overlapping subsets of users $\mcS_A$ and $\mcS_B$ (i.e., $\mcS_A, \mcS_B \subset \mcU$ and $\mcS_A \cap \mcS_B = \emptyset$) and apply models $A$ and $B$ to each subset.
Let $n = |L_u^A| = |L_u^B|$ be the size of the recommendation list, which we assume to be constant.
The subset average causal effect is then defined as
\begin{equation}
	\hat{\tau}_{A} = \frac{1}{n |\mcS_A|} \sum_{u \in \mcS_A} \sum_{i \in L_u^A} \tau_{ui} .
\end{equation}
This converges to $\tau_{A}$ as $|\mcS_A|$ increases.

The typical evaluation metrics for A/B testing are either based on total user interactions (such as sales or user engagement) or only on interactions with recommended lists (such as click-through rates or conversion rates)~\cite{Jannach19}.
Here we show that the former is a valid evaluation for the causal effect.
The total user interactions divided by the number of recommendations can be expressed as
\begin{align} 
	\label{eq:AB_total}
	\hat{Y}_A^{\textnormal{total}} 
	&= \frac{1}{ n |\mcS_A| } \sum_{u \in \mcS_A} \sum_{i \in \mcI} Y_{ui} \nonumber\\
	&= \frac{1}{ n |\mcS_A| } \sum_{u \in \mcS_A} \left(\sum_{i \in L_u^A} \YT + \sum_{i \in \mcI \backslash L_u^A} \YC \right)
	= \frac{1}{ n |\mcS_A| } \sum_{u \in \mcS_A} \left(\sum_{i \in L_u^A} (\tau_{ui} + \YC) + \sum_{i \in \mcI \backslash L_u^A} \YC \right) \nonumber\\
	&= \hat{\tau}_{A} + \frac{1}{ n |\mcS_A| } \sum_{u \in \mcS_A} \sum_{i \in \mcI} \YC.
\end{align}
Because the rightmost term in the final equation does not depend on the model, we can compare $\hat{\tau}_{A}$ and $\hat{\tau}_{B}$ by comparing $\hat{Y}_A^{\textnormal{total}}$ and $\hat{Y}_B^{\textnormal{total}}$.
On the other hand, the average interactions with the recommended lists can be expressed as
\begin{align} 
	\label{eq:AB_list}
	\hat{Y}_A^{\textnormal{list}} 
	&= \frac{1}{n |\mcS_A|} \sum_{u \in \mcS_A} \sum_{i \in L_u^A} Y_{ui} \nonumber\\
	&= \frac{1}{n |\mcS_A|} \sum_{u \in \mcS_A} \sum_{i \in L_u^A} \YT \neq \hat{\tau}_{A}.
\end{align}
Hence, the evaluation based only on interactions with recommended lists is not valid testing for the causal effect.

Although A/B testing with Eq. (\ref{eq:AB_total}) can be used for unbiased model comparisons, it may have large variance due to the variability in natural user behaviors (i.e., the potential outcomes under no recommendations, $\YC$).
If users in $\mcS_A$ tend to purchase more items than those in $\mcS_B$, $\sum_{u \in \mcS_A} \sum_{i \in \mcI} \YC$ becomes larger than $\sum_{u \in \mcS_B} \sum_{i \in \mcI} \YC$, thereby altering the comparison in Eq. (\ref{eq:AB_total}).
To minimize such discrepancies, a sufficiently large number of users need to be recruited for A/B testing.
We thus introduce more efficient evaluation methods in the next subsection.

\subsection{Interleaving for the Causal Effect}
In this subsection, we propose interleaving methods for the online evaluation of the causal effects of recommendations.
Previous interleaving methods only measure outcomes in the interleaved lists: they only include $\YT$ and lack information on $\YC$.
Further, if the item selection for the interleaved list is not randomized controlled, the naive estimate from the observed outcomes might be biased due to the confounding between recommendations and potential outcomes.
We need to remedy the bias for valid comparison.

Here we describe the problem setting of interleaving for the causal effect.
For each user $u$, we construct the interleaved list $L_u$ from the compared lists $L_u^A$ and $L_u^B$.
We observe outcomes $\{Y_{ui}\}$ for all items $i \in \mcI$.
Note that $Y_{ui} = \YT$ if item $i$ is in the interleaved list ($i \in L_u$ or equivalently, $Z_{ui} = 1$) and $Y_{ui} = \YC$ if it is not in the list ($i \in \mcI \backslash L_u$ or equivalently, $Z_{ui} = 0$).
We want to compare the average causal effects of lists $L_u^A$ and $L_u^B$:
\begin{equation}
	\tau_{L_u^A} = \frac{1}{n} \sum_{i \in L_u^A} \tau_{ui}, \quad
	\tau_{L_u^B} = \frac{1}{n} \sum_{i \in L_u^B} \tau_{ui}.
\end{equation}
We need to estimate the above values from observed outcomes because we cannot directly observe $\tau_{ui}$.

If the items in $L_u^A$ and $L_u^B$ are randomly assigned to the interleaved list independent of the potential outcomes, that is, $\left( \YT, \YC \right) \perp Z_{ui}$, the case can be regarded as a randomized controlled trial (RCT)~\cite{Rubin74,Imbens15}.\footnote{For our interleaving methods, the independence is required only for the items in the union of $L_u^A$ and $L_u^B$.}
We can then simply estimate $\tau_{L_u^A}$ as the difference in average outcomes for items on and not on the interleaved list:
\begin{equation}
	\label{eq:RCT}
	\left( \hat{\tau}_{L_u^A} \right)_{\textnormal{RCT}} = \frac{1}{|L_u^A \cap L_u|} \sum_{i \in L_u^A \cap L_u} Y_{ui} - \frac{1}{|L_u^A \backslash L_u|} \sum_{i \in L_u^A \backslash L_u} Y_{ui}.
\end{equation}
One way to realize such a randomized assignment is to select $n$ items from $L_u^A \cup L_u^B$ with equal probability: $p = n/ |L_u^A \cup L_u^B|$.
We call this method equal probability interleaving (EPI).

The independence requirement heavily restricts the potential design space of interleaving methods.
We thus derive estimates that are applicable to more general cases.
Denote the probability (also called the propensity) of being included in the interleaved list $L_u$ by $p_{ui} = \E[Z_{ui}=1|X_{ui}]$.
We assume that 1) the covariates $X_{ui}$ contain all confounders of $\left( \YT, \YC \right)$ and $Z_{ui}$, and 2) the treatment assignment is not deterministic ($0 < p_{ui} < 1$ for $i \in L_u^A \cup L_u^B$).\footnote{Taken together, these two assumptions are called \textit{strongly ignorable treatment assignment}~\cite{Rosenbaum83}.}
Assumption 1 is equivalent to conditional independence: $\left( \YT, \YC \right) \perp Z_{ui} | X_{ui}$.
When we design an interleaving method, we know the covariates that affect $Z_{ui}$ and Assumption 1 can always be satisfied.\footnote{Confounders are covariates that affect both $\left( \YT, \YC \right)$ and $Z_{ui}$, and they are subsets of covariates that affect $Z_{ui}$. Hence, including the latter in $X_{ui}$ is a sufficient condition for Assumption 1.}
Therefore, the only restriction for interleaving methods is Assumption 2 (also called \textit{positivity}).
 
Under these assumptions, we can construct an unbiased estimator using IPS weighting~\cite{Lunceford04}:
\begin{equation}
	\label{eq:IPS}
	\left( \hat{\tau}_{L_u^A} \right)_{\textnormal{IPS}} = \frac{1}{n} \sum_{i \in L_u^A \cap L_u} \frac{Y_{ui}}{p_{ui}} - \frac{1}{n} \sum_{i \in L_u^A \backslash L_u} \frac{Y_{ui}}{1-p_{ui}}
	=  \frac{1}{n} \sum_{i \in L_u^A} \left( \frac{Z_{ui} Y_{ui}}{p_{ui}} -  \frac{(1-Z_{ui})Y_{ui}}{1-p_{ui}} \right).
\end{equation}
This estimator is unbiased since
\begin{align}
	\E\left[\frac{Z_{ui} Y_{ui}}{p_{ui}} - \frac{(1-Z_{ui})Y_{ui}}{1-p_{ui}} \middle| X_{ui} \right] 
	& = \E\left[\frac{Z_{ui}\YT}{p_{ui}} - \frac{(1-Z_{ui})\YC}{1-p_{ui}} \middle| X_{ui} \right] 
	  =\frac{\E[Z_{ui}| X_{ui}] \YT}{p_{ui}} - \frac{\E[(1-Z_{ui})| X_{ui}] \YC}{1-p_{ui}}  \nonumber\\
	& = \frac{p_{ui}\YT}{p_{ui}} - \frac{(1-p_{ui}) \YC}{1-p_{ui}} = \tau_{ui}.
\end{align}

We propose a general framework for interleaving as follows.
\begin{enumerate}
	\item Construct interleaved lists $\{L_u\}$ using an interleaving method that satisfies positivity (Assumption 2).
	\item Conduct online experiments and obtain outcomes $\{Y_{ui}\}$.
	\item Estimate $\tau_{L_u^A}$ and $\tau_{L_u^B}$ by Eq. (\ref{eq:IPS}) and compare them.
\end{enumerate}

\begin{algorithm}[btp] 
	\caption{Causal Balanced Interleaving ({\textit{CBI}}).}
	\label{algo:CBI}
	\DontPrintSemicolon
	\KwIn{Compared lists $L_u^A$ and $L_u^B$}
	\KwOut{Interleaved list $L_u$ with size $n$ ($n=|L_u^A|=|L_u^B|$)} 
	$L_u \gets ()$ \tcp*[r]{initialize interleaved list}
	$ANext \gets RandBit()=1$ \tcp*[r]{Randomly select whether starting from A or B}
	\While{$|L_u| < n$}{
		\eIf{$ANext$}{
			$L_u \gets L_u + RandomChoiceFrom(L_u^A \backslash L_u)$
			\tcp*[r]{Randomly choose one item from $L_u^A$ not yet in $L_u$}
			$ANext \gets False$
		}{
			$L_u \gets L_u + RandomChoiceFrom(L_u^B \backslash L_u$)
			\tcp*[r]{Randomly choose one item from $L_u^B$ not yet in $L_u$}
			$ANext \gets True$
		}
	}
	\Return{$L_u$}\;
\end{algorithm}

As an example of a valid interleaving method that satisfies positivity, we propose causal balanced interleaving (CBI), the pseudo-code for which is shown in Algorithm~\ref{algo:CBI}.
CBI alternatively selects items from each list to balance the items chosen from each list.
The item choice in each round is not deterministic in order to satisfy the positivity required for causal effect estimates.
The propensity depends on whether an item is in the intersection, $\boldsymbol{1} (i \in L_u^A \cap L_u^B)$.
If an item is included in both lists, it has a greater probability of being chosen.
The propensity also depends on the cardinality of the union of the compared lists, $|L_u^A \cup L_u^B|$, because smaller cardinality implies that each item has a greater chance of being selected.
The possible values of the covariates are limited: $\boldsymbol{1} (i \in L_u^A \cap L_u^B)$ is binary and $|L_u^A \cup L_u^B| \in [n, 2n]$.
Hence, we can easily compute the propensity numerically by repeating Algorithm~\ref{algo:CBI} a sufficient number of times and recording $Z_{ui}$ for each combination of covariates.

\section{Experiments}
\subsection{Experimental Setup}
We experimented with the following online evaluation methods.\footnote{For reproducibility, the code is available at https://github.com/masatoh73/causal-interleaving.}
\begin{itemize}
	\item AB-total: A/B testing evaluated by the total user interactions, as expressed in Eq. (\ref{eq:AB_total}). 
	\item AB-list: A/B testing evaluated by user interactions only with items on the recommended list, as in Eq. (\ref{eq:AB_list}).
	\item EPI-RCT: Interleaving to select items from $L_u^A \cup L_u^B$ with equal probability and evaluation using Eq. (\ref{eq:RCT}).
	\item CBI-RCT: Interleaving by Algorithm \ref{algo:CBI} and evaluation using Eq. (\ref{eq:RCT}), that is, no bias correction by IPS.
	\item CBI-IPS: Interleaving by Algorithm \ref{algo:CBI} and evaluation using Eq. (\ref{eq:IPS}).
\end{itemize}
Through the experiments, we aim to answer the following research questions: RQ1) Which method produces valid (unbiased) estimates of the true differences in average causal effects (\ref{sec:validity})?, and RQ2) Are the proposed interleaving methods more efficient (do they require fewer experimental users) than AB testing (\ref{sec:efficiency})?
We first prepared semi-synthetic datasets that contain both potential outcomes $\YT$ and $\YC$ for all user-item pairs.
Because we observe $Y_{ui} =\YT$ if $Z_{ui} = 1$ and $Y_{ui} =\YC$ if $Z_{ui} = 0$, both potential outcomes are necessary to simulate user outcomes under various ranking models and online evaluation methods.
Following the procedure described in~\cite{Sato21}, we generated two datasets: one is based on the Dunnhumby dataset,\footnote{https://www.dunnhumby.com/careers/engineering/sourcefiles} and the other is based on the MovieLens-1M (ML-1M) dataset~\cite{Harper15}.\footnote{https://grouplens.org/datasets/movielens}
The detail and rationale of ML one are described in Section 5.1 of~\cite{Sato21} and that of DH one are described in 5.1.1 of~\cite{Sato20}.
Each dataset is comprised of independently generated training and testing data.
The testing data were used to simulate online evaluation, and the training data were used to train the following models:\footnote{We used hyper-parameters for CP@10, described in the ancillary files at http://arxiv.org/abs/2012.09442.} the causality-aware user-based neighborhood methods (CUBN) with outcome similarity (-O) and treatment similarity (-T)~\cite{Sato21}, the uplift-based pointwise and pairwise learning methods (ULRMF and ULBPR)~\cite{Sato19}, the Bayesian personalized ranking method (BPR)~\cite{Rendle09}, and the user-based neighborhood method (UBN)~\cite{Ning15}.
We compared two models among {CUBN-T, ULRMF, BPR} on the Dunnhumby data and two models among {CUBN-O, ULBPR, UBN} on the ML-1M data.\footnote{We intended to compare models of different families, i.e., one of \{CUBN-T, CUBN-O\} with one of \{ULBPR, ULRMF\}.}
The average causal effect $\overline{\tau_{L_u^{model}}}$ and the average treated outcomes $\overline{Y_{L_u^{model}}^{\textnormal{T}}}$ of the trained models are listed in Table \ref{tab:stat_models}.
The superior models in terms of the average causal effect do not necessarily have higher average treated outcomes.
That is, we may mistakenly select a poor model in terms of the causal effect if we only evaluate the outcomes of the recommended items.

\begin{table}[htbp]
	\caption{Averages of causal effect and potential outcomes under treatment with recommendation lists of size $n=10$. }
	\label{tab:stat_models}
	\centering
	\small
	\begin{tabular}{lccc|ccc}
		\toprule
		& \multicolumn{3}{c}{Dunnhumby-Original} & \multicolumn{3}{c}{MovieLens-1M} \\
		& CUBN-T & ULRMF & BPR & CUBN-O & ULBPR & UBN  \\
		\midrule
		$\overline{\tau_{L_u^{model}}}$ & 0.0507 & 0.0347 & 0.0295 
		& 0.332 & 0.280 & -0.186\\
		$\overline{Y_{L_u^{model}}^{\textnormal{T}}}$ & 0.1359 & 0.1396 & 0.1869 &
		0.341 & 0.285 & 0.308 \\
		\bottomrule
	\end{tabular}
\end{table}

Our protocol for simulating online experiments is the following.
First, we randomly select a subset of users and generate lists $L_u^A, L_u^B$ using compared models.
For the A/B testing methods (AB-total, AB-list), we further split the subset into two groups: $\mcS_A$ and $\mcS_B$, and $\{L_u^A\}$ and $\{L_u^B\}$ are recommended for each group, respectively.
For the interleaving methods (EPI-RCT, CBI-RCT, CBI-IPS), we generate interleaved recommendation lists using EPI or CBI.
In the simulation, \textit{recommendation} means that $Z_{ui}$ is set to $1$, and user outcomes $\{Y_{ui}\}$ are \textit{observed} by calculating $Y_{ui} = Z_{ui}\YT + (1-Z_{ui})\YC$ with potential outcomes $\YT$ and $\YC$.
Using the observed outcomes, we estimate the difference in the average causal effects of the compared models: $\tau_{A} - \tau_{B}$.
We repeated the above protocol $10,000$ times and recorded the estimated differences using each online evaluation method.
The size of recommendation list was set to 10.

\subsection{Results and Discussion}

\subsubsection{Validity of the evaluation methods}
\label{sec:validity}
We evaluated the validity of the online evaluation methods using random subsets of 1,000 users.
The means and standard deviations of the estimated differences are shown in Table \ref{tab:estimated_difference}.
The means obtained by EPI-RCT and CBI-IPS are close to the true differences.
The means obtained by AB-total are also close to the true value for Dunnhumby but deviate slightly for ML-1M.
The AB-list often yields estimates that differ substantially from the true values but are similar to the differences in treated outcomes, $\overline{Y_{L_u^{model}}^{\textnormal{T}}}$, as shown in Table \ref{tab:stat_models}.
This is expected because the AB-list evaluates $\YT$, not $\tau_{ui}$, as expressed in Eq. (\ref{eq:AB_list}).
Further, the CBI-RCT estimates also deviate from the true differences in most cases.\footnote{In the comparisons of CUBN-O \& UBN and ULBPR \& UBN, the results of CBI-RCT and CBI-IPS are identical. There was no overlaps between $L_u^A$ and $L_u^B$ in these comparisons, and the propensity was constant. Hence, IPS was not necessary, and  CBI-RCT and CBI-IPS were equivalent.}
This is due to the bias induced by the uneven probability of recommendation in interleaving.
Conversely, CBI-IPS successfully removes the bias and produces estimates centered around the true values.

\begin{table}[htbp]
	\caption{Estimated differences between the causal effects of the compared models (mean $\pm$ standard deviations for 10,000 simulated runs). The results highlighted in bold indicate that the true values are within the 95\% confidence intervals of the mean estimates.}
	\label{tab:estimated_difference}
	\centering
	\small 
	\begin{tabular}{llll|lll}
		\toprule
		& \multicolumn{3}{c}{Dunnhumby-Original} & \multicolumn{3}{c}{MovieLens-1M} \\
		& CUBN-T \& BPR  & CUBN-T \& ULRMF  &  ULRMF \& BPR 
		& CUBN-O \& UBN  & CUBN-O \& ULBPR  &  ULBPR \& UBN  \\
		\midrule
		Truth & 0.0212 & 0.0160 & 0.0052 & 
		0.5177 & 0.0512 & 0.4665 \\
		AB-total & \textbf{0.0210} $\pm$ 0.0399 & \textbf{0.0159} $\pm$ 0.0399 & \textbf{0.0051} $\pm$ 0.0397 & 
		\textbf{0.5301} $\pm$ 1.2048 & \textbf{0.0635} $\pm$ 1.2102 & \textbf{0.4789} $\pm$ 1.2052 \\
		AB-list & -0.0510 $\pm$ 0.0071 & -0.0037 $\pm$ 0.0065 & -0.0471 $\pm$ 0.0073 & 
		0.0325 $\pm$ 0.0104 & 0.0550 $\pm$ 0.0104 & -0.0226 $\pm$ 0.0100 \\
		EPI-RCT & \textbf{0.0212} $\pm$ 0.0069 & \textbf{0.0159} $\pm$ 0.0075 & \textbf{0.0053} $\pm$ 0.0076 & 
		\textbf{0.5178} $\pm$ 0.0137 & \textbf{0.0512} $\pm$ 0.0083 & \textbf{0.4666} $\pm$ 0.0135 \\
		CBI-RCT & 0.0429 $\pm$ 0.0067 & 0.0192 $\pm$ 0.0067 & 0.0188 $\pm$ 0.0076 & 
		\textbf{0.5179} $\pm$ 0.0126 & 0.0444 $\pm$ 0.0066 & \textbf{0.4667} $\pm$ 0.0126 \\
		CBI-IPS & \textbf{0.0213} $\pm$ 0.0063 & \textbf{0.0160} $\pm$ 0.0066 & \textbf{0.0051} $\pm$ 0.0070 & 
		\textbf{0.5179} $\pm$ 0.0126 & \textbf{0.0512} $\pm$ 0.0075 & \textbf{0.4667} $\pm$ 0.0126 \\
		\bottomrule
	\end{tabular}
\end{table}

\subsubsection{Efficiency of the interleaving methods}
\label{sec:efficiency}
\begin{figure*}[htbp]
	\begin{center}
		\subfigure[CUBN-T \& BPR in Dunnhumby.]{\includegraphics[width=0.32\textwidth]{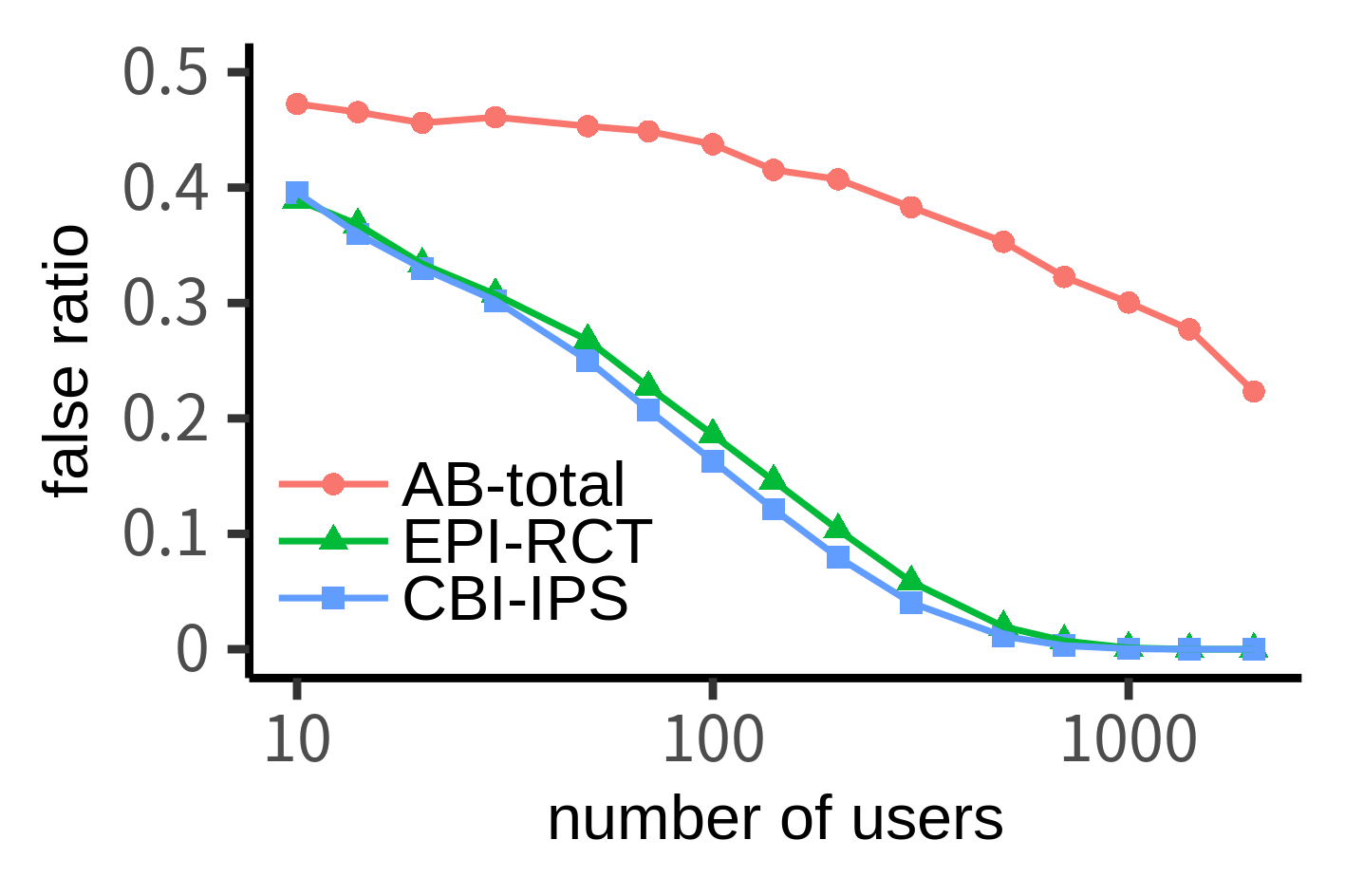}}
		\subfigure[CUBN-T \& ULRMF in Dunnhumby.]{\includegraphics[width=0.32\textwidth]{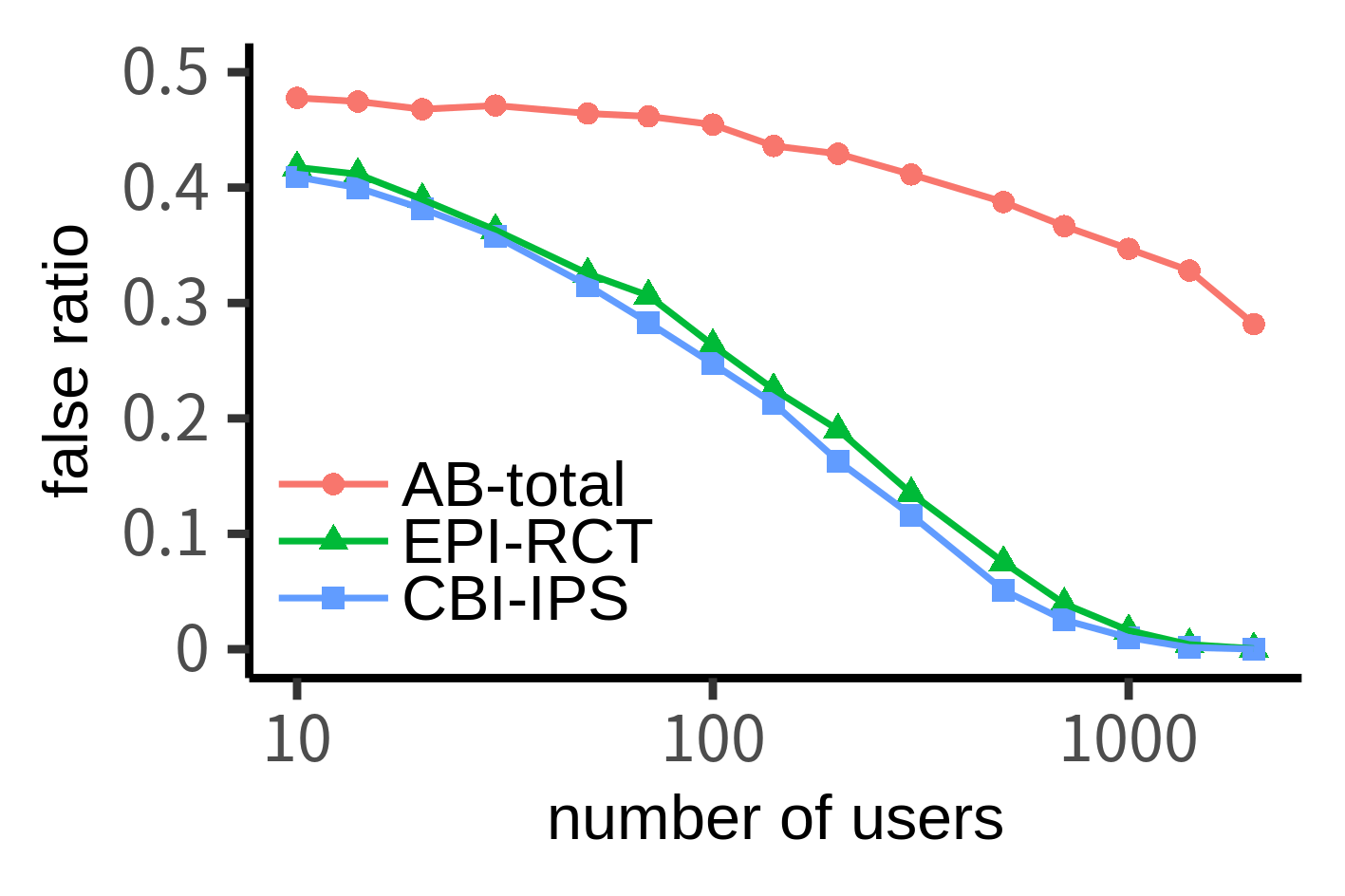}}
		\subfigure[ULRMF \& BPR in Dunnhumby.]{\includegraphics[width=0.32\textwidth]{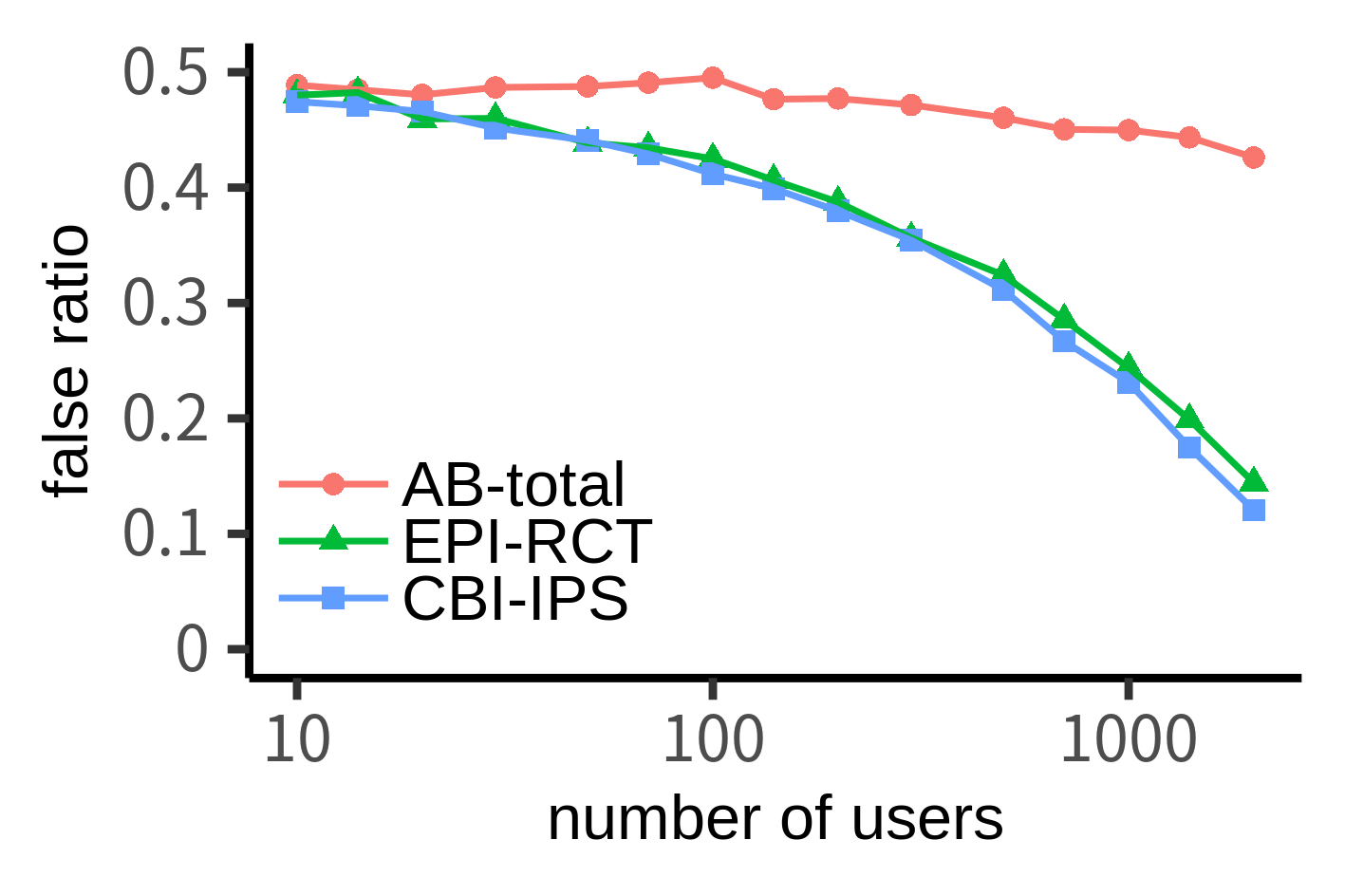}}
		\subfigure[CUBN-O \& UBN in ML-1M.]{\includegraphics[width=0.32\textwidth]{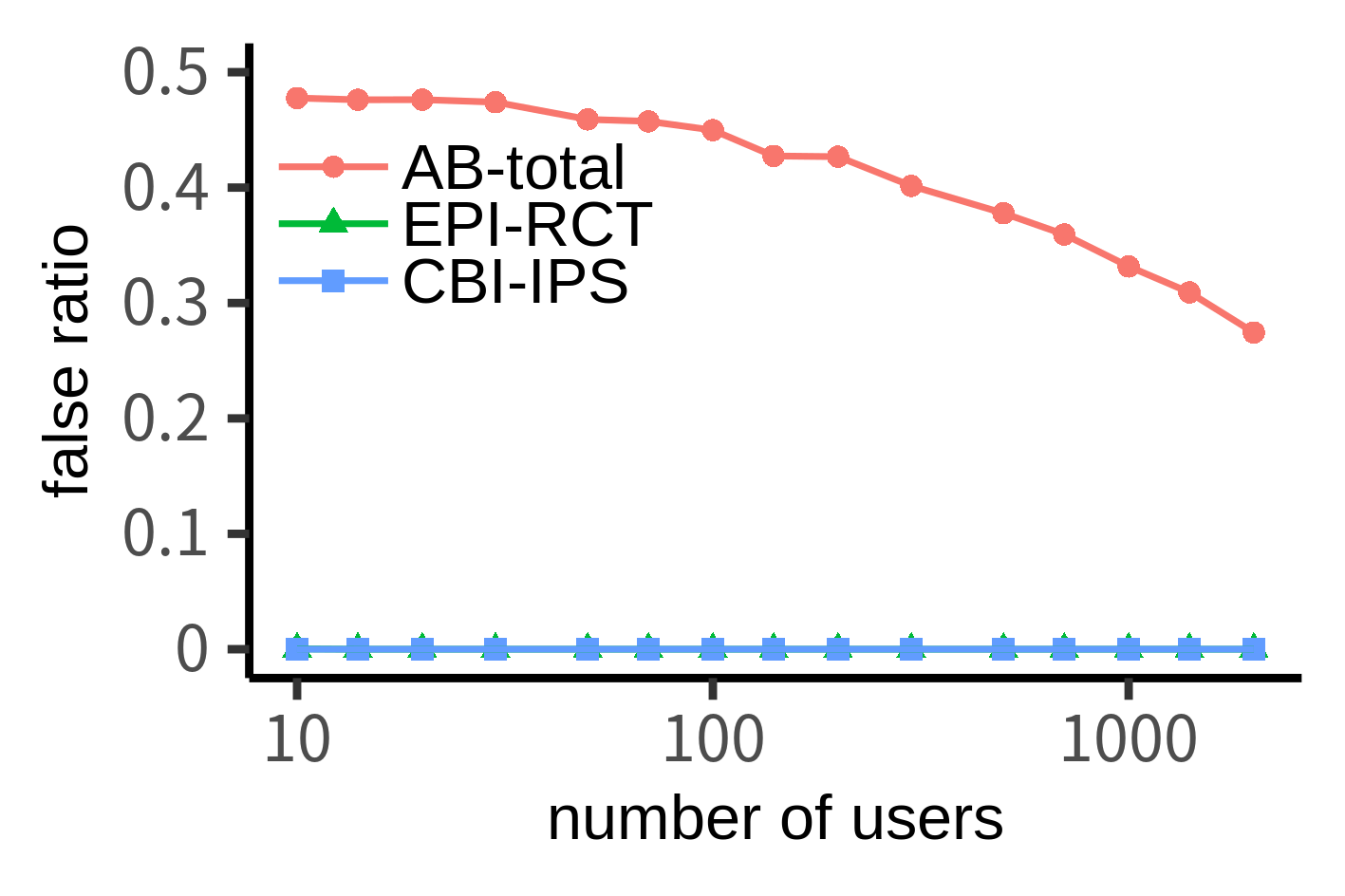}}
		\subfigure[CUBN-O \& ULBPR in ML-1M.]{\includegraphics[width=0.32\textwidth]{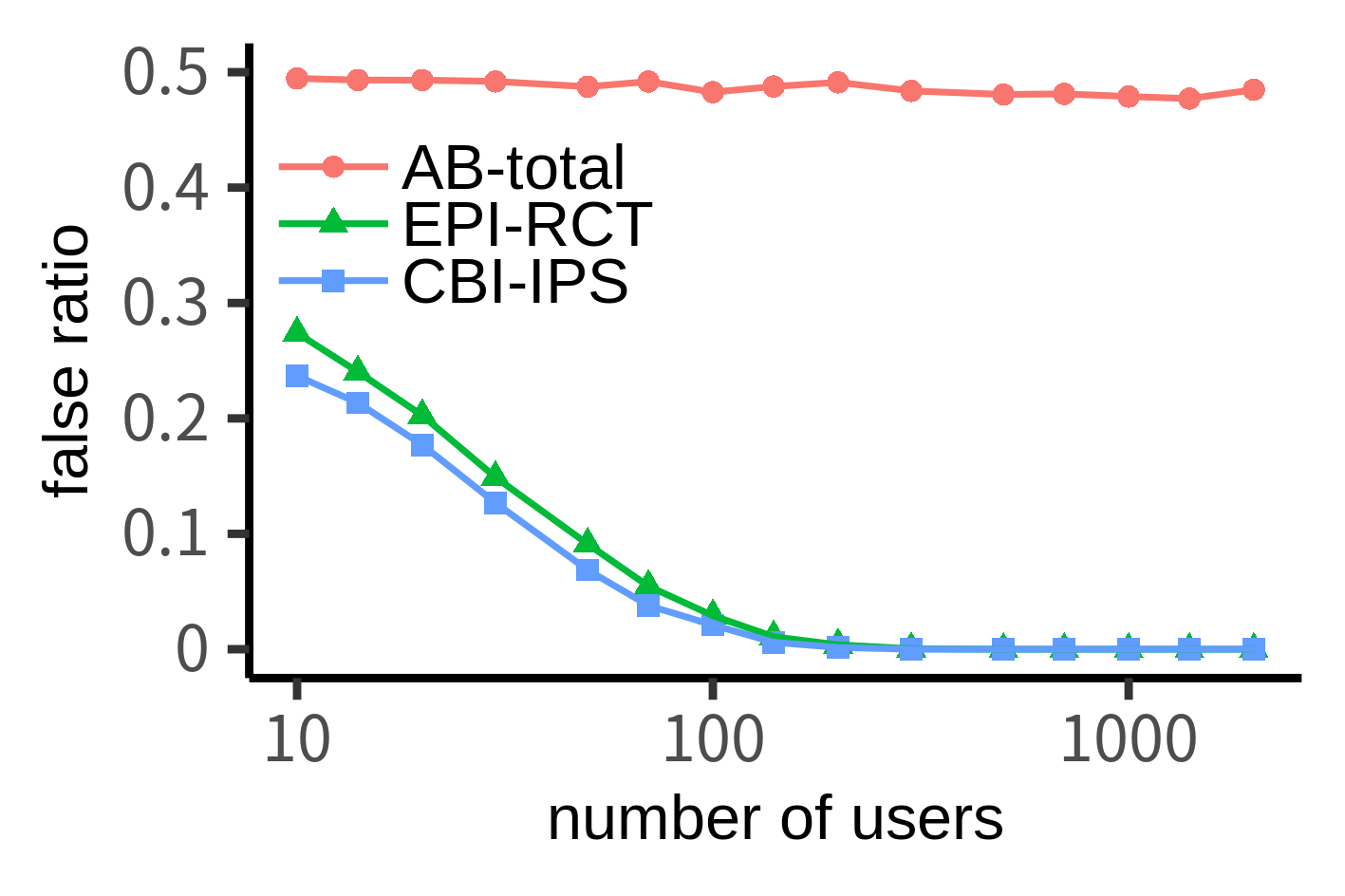}}
		\subfigure[ULBPR \& UBN in ML-1M.]{\includegraphics[width=0.32\textwidth]{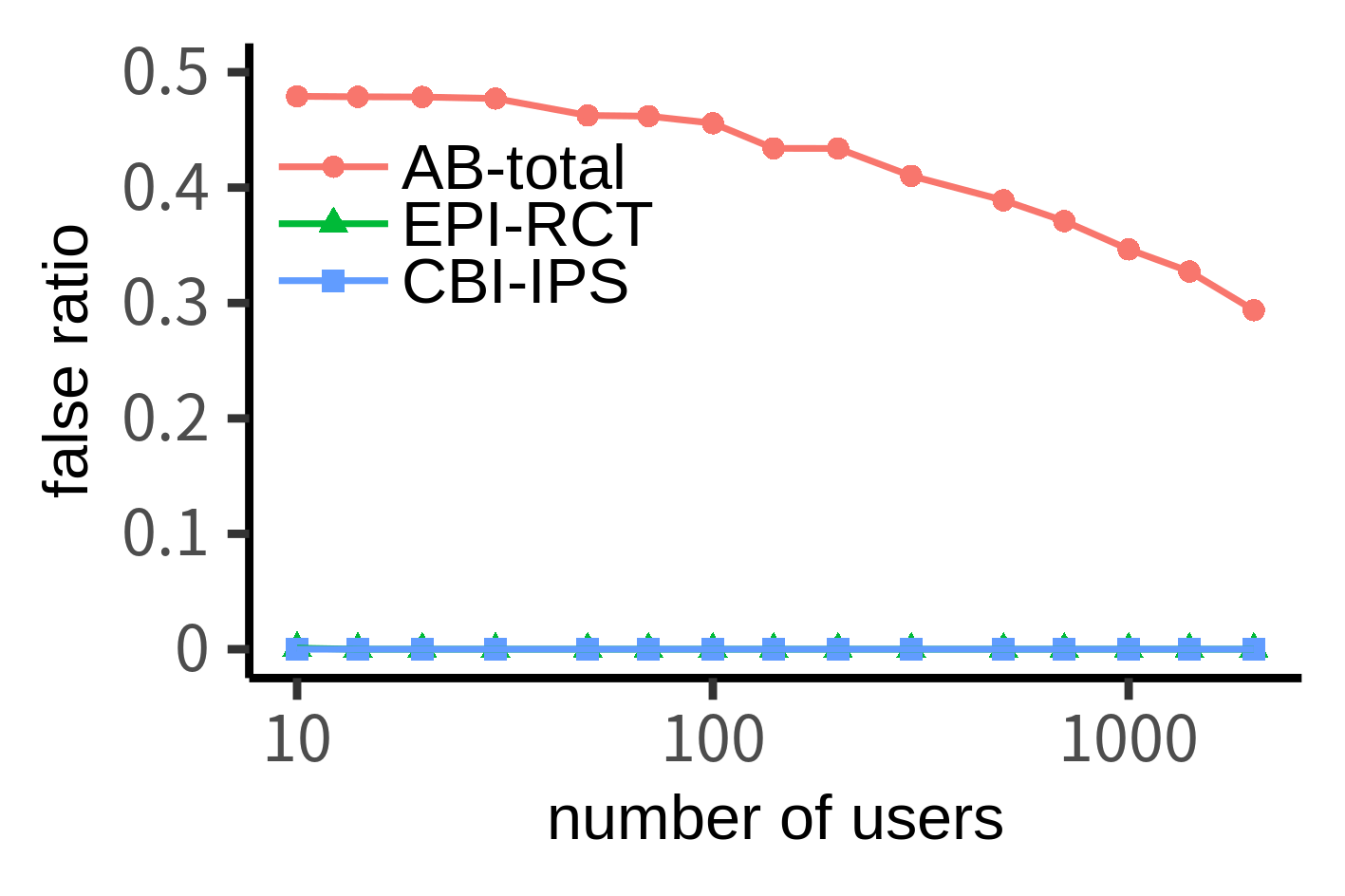}}
		\caption{Dependence on the number of users.}
		\label{fig:efficiency}
	\end{center}
	\Description[The false ratios of the proposed interleaving methods decrease with smaller numbers of users compared to the AB testing method.]{In Dunnhumby, the false ratios of AB-total in 2,000 users were about the same with the false ratios of EPI-RCT and CBI-IPS in 70 users. In Movielens-1M, the false ratios of AB-total in 2,000 users were much worse than the false ratios of EPI-RCT and CBI-IPS in 10 users.}
\end{figure*}

We compared the efficiency of AB-total, EPI-RCT, and CBI-IPS, all of which were shown to be valid in the previous section.
We simulated user subsets of various sizes in \{10, 14, 20, 30, 50, 70, 100, 140, 200, 300, 500, 700, 1000, 1400, 2000\} and evaluated the ratio of false judgments (when the sign of the estimated difference is the opposite of the truth).
Figure \ref{fig:efficiency} shows the ratio of false judgments according to the number of users.
As the number of users increases, the false ratios of CBI-IPS and EPI-RCT decrease more rapidly than that of AB-total does.
For the Dunnhumby dataset, AB-total requires around 30 times more users than CBI-IPS and EPI-RCT to achieve the same false ratio.
For the ML-1M dataset, AB-total did not reach the same false ratio in the experimental range of subset sizes.
These results demonstrate the superior efficiency of the proposed interleaving methods.
Furthermore, CBI-IPS tends to be slightly more efficient than EPI-RCT, as expected from the smaller standard deviations shown in Table \ref{tab:estimated_difference}.
This is probably because the number of items selected from the compared lists is balanced in this interleaving method.

\section{Conclusions}
In this paper, we proposed the first interleaving methods for comparing recommender models in terms of causal effects.
To realize unbiased model comparisons, our methods either select items with equal probabilities or weight the outcomes using IPS.
We simulated online experiments and verified that our interleaving methods and an A/B testing method are unbiased and that our interleaving methods are largely more efficient than the A/B testing method.
In the future, we plan to extend our methods to multileaving.
Online experimentation in real recommendation services will also be important for future work.

\bibliographystyle{ACM-Reference-Format}
\bibliography{ref}


\begin{thebibliography}{33}


\ifx \showCODEN    \undefined \def \showCODEN     #1{\unskip}     \fi
\ifx \showDOI      \undefined \def \showDOI       #1{#1}\fi
\ifx \showISBNx    \undefined \def \showISBNx     #1{\unskip}     \fi
\ifx \showISBNxiii \undefined \def \showISBNxiii  #1{\unskip}     \fi
\ifx \showISSN     \undefined \def \showISSN      #1{\unskip}     \fi
\ifx \showLCCN     \undefined \def \showLCCN      #1{\unskip}     \fi
\ifx \shownote     \undefined \def \shownote      #1{#1}          \fi
\ifx \showarticletitle \undefined \def \showarticletitle #1{#1}   \fi
\ifx \showURL      \undefined \def \showURL       {\relax}        \fi
\providecommand\bibfield[2]{#2}
\providecommand\bibinfo[2]{#2}
\providecommand\natexlab[1]{#1}
\providecommand\showeprint[2][]{arXiv:#2}

\bibitem[\protect\citeauthoryear{Bodapati}{Bodapati}{2008}]%
        {Bodapati08}
\bibfield{author}{\bibinfo{person}{Anand~V Bodapati}.}
  \bibinfo{year}{2008}\natexlab{}.
\newblock \showarticletitle{Recommendation systems with purchase data}.
\newblock \bibinfo{journal}{\emph{Journal of marketing research}}
  \bibinfo{volume}{45}, \bibinfo{number}{1} (\bibinfo{year}{2008}),
  \bibinfo{pages}{77--93}.
\newblock


\bibitem[\protect\citeauthoryear{Bonner and Vasile}{Bonner and Vasile}{2018}]%
        {Bonner18}
\bibfield{author}{\bibinfo{person}{Stephen Bonner} {and}
  \bibinfo{person}{Flavian Vasile}.} \bibinfo{year}{2018}\natexlab{}.
\newblock \showarticletitle{Causal Embeddings for Recommendation}. In
  \bibinfo{booktitle}{\emph{Proceedings of the 12th ACM Conference on
  Recommender Systems}} (Vancouver, British Columbia, Canada)
  \emph{(\bibinfo{series}{RecSys ’18})}. \bibinfo{publisher}{Association for
  Computing Machinery}, \bibinfo{address}{New York, NY, USA},
  \bibinfo{pages}{104–112}.
\newblock
\showISBNx{9781450359016}
\urldef\tempurl%
\url{https://doi.org/10.1145/3240323.3240360}
\showDOI{\tempurl}


\bibitem[\protect\citeauthoryear{Chapelle, Joachims, Radlinski, and
  Yue}{Chapelle et~al\mbox{.}}{2012}]%
        {Chapelle12}
\bibfield{author}{\bibinfo{person}{Olivier Chapelle}, \bibinfo{person}{Thorsten
  Joachims}, \bibinfo{person}{Filip Radlinski}, {and} \bibinfo{person}{Yisong
  Yue}.} \bibinfo{year}{2012}\natexlab{}.
\newblock \showarticletitle{Large-Scale Validation and Analysis of Interleaved
  Search Evaluation}.
\newblock \bibinfo{journal}{\emph{ACM Trans. Inf. Syst.}} \bibinfo{volume}{30},
  \bibinfo{number}{1}, Article \bibinfo{articleno}{6} (\bibinfo{date}{March}
  \bibinfo{year}{2012}), \bibinfo{numpages}{41}~pages.
\newblock
\showISSN{1046-8188}
\urldef\tempurl%
\url{https://doi.org/10.1145/2094072.2094078}
\showDOI{\tempurl}


\bibitem[\protect\citeauthoryear{Chen, Dong, Wang, Feng, Wang, and He}{Chen
  et~al\mbox{.}}{2020}]%
        {Chen20}
\bibfield{author}{\bibinfo{person}{Jiawei Chen}, \bibinfo{person}{Hande Dong},
  \bibinfo{person}{Xiang Wang}, \bibinfo{person}{Fuli Feng},
  \bibinfo{person}{Meng Wang}, {and} \bibinfo{person}{Xiangnan He}.}
  \bibinfo{year}{2020}\natexlab{}.
\newblock \showarticletitle{Bias and debias in recommender system: A survey and
  future directions}.
\newblock \bibinfo{journal}{\emph{arXiv preprint arXiv:2010.03240}}
  (\bibinfo{year}{2020}).
\newblock


\bibitem[\protect\citeauthoryear{Cosley, Lam, Albert, Konstan, and
  Riedl}{Cosley et~al\mbox{.}}{2003}]%
        {Cosley03}
\bibfield{author}{\bibinfo{person}{Dan Cosley}, \bibinfo{person}{Shyong~K.
  Lam}, \bibinfo{person}{Istvan Albert}, \bibinfo{person}{Joseph~A. Konstan},
  {and} \bibinfo{person}{John Riedl}.} \bibinfo{year}{2003}\natexlab{}.
\newblock \showarticletitle{Is Seeing Believing? How Recommender System
  Interfaces Affect Users' Opinions}. In \bibinfo{booktitle}{\emph{Proceedings
  of the SIGCHI Conference on Human Factors in Computing Systems}} (Ft.
  Lauderdale, Florida, USA) \emph{(\bibinfo{series}{CHI '03})}.
  \bibinfo{publisher}{Association for Computing Machinery},
  \bibinfo{address}{New York, NY, USA}, \bibinfo{pages}{585–592}.
\newblock
\showISBNx{1581136307}
\urldef\tempurl%
\url{https://doi.org/10.1145/642611.642713}
\showDOI{\tempurl}


\bibitem[\protect\citeauthoryear{Dias, Locher, Li, El-Deredy, and Lisboa}{Dias
  et~al\mbox{.}}{2008}]%
        {Dias08}
\bibfield{author}{\bibinfo{person}{M.~Benjamin Dias},
  \bibinfo{person}{Dominique Locher}, \bibinfo{person}{Ming Li},
  \bibinfo{person}{Wael El-Deredy}, {and} \bibinfo{person}{Paulo~J.G. Lisboa}.}
  \bibinfo{year}{2008}\natexlab{}.
\newblock \showarticletitle{The Value of Personalised Recommender Systems to
  E-Business: A Case Study}. In \bibinfo{booktitle}{\emph{Proceedings of the
  2008 ACM Conference on Recommender Systems}} (Lausanne, Switzerland)
  \emph{(\bibinfo{series}{RecSys ’08})}. \bibinfo{publisher}{Association for
  Computing Machinery}, \bibinfo{address}{New York, NY, USA},
  \bibinfo{pages}{291–294}.
\newblock
\showISBNx{9781605580937}
\urldef\tempurl%
\url{https://doi.org/10.1145/1454008.1454054}
\showDOI{\tempurl}


\bibitem[\protect\citeauthoryear{Harper and Konstan}{Harper and
  Konstan}{2015}]%
        {Harper15}
\bibfield{author}{\bibinfo{person}{F.~Maxwell Harper} {and}
  \bibinfo{person}{Joseph~A. Konstan}.} \bibinfo{year}{2015}\natexlab{}.
\newblock \showarticletitle{The MovieLens Datasets: History and Context}.
\newblock \bibinfo{journal}{\emph{ACM Trans. Interact. Intell. Syst.}}
  \bibinfo{volume}{5}, \bibinfo{number}{4}, Article \bibinfo{articleno}{19}
  (\bibinfo{date}{Dec.} \bibinfo{year}{2015}), \bibinfo{numpages}{19}~pages.
\newblock
\showISSN{2160-6455}
\urldef\tempurl%
\url{https://doi.org/10.1145/2827872}
\showDOI{\tempurl}


\bibitem[\protect\citeauthoryear{Hern{\'a}n and Robins}{Hern{\'a}n and
  Robins}{2020}]%
        {Hernan20}
\bibfield{author}{\bibinfo{person}{MA Hern{\'a}n} {and} \bibinfo{person}{JM
  Robins}.} \bibinfo{year}{2020}\natexlab{}.
\newblock \showarticletitle{Causal inference: What if}.
\newblock \bibinfo{journal}{\emph{Boca Raton: Chapman \& Hill/CRC}}
  (\bibinfo{year}{2020}).
\newblock


\bibitem[\protect\citeauthoryear{Hofmann, Whiteson, and de~Rijke}{Hofmann
  et~al\mbox{.}}{2011}]%
        {Hofmann11}
\bibfield{author}{\bibinfo{person}{Katja Hofmann}, \bibinfo{person}{Shimon
  Whiteson}, {and} \bibinfo{person}{Maarten de Rijke}.}
  \bibinfo{year}{2011}\natexlab{}.
\newblock \showarticletitle{A Probabilistic Method for Inferring Preferences
  from Clicks}. In \bibinfo{booktitle}{\emph{Proceedings of the 20th ACM
  International Conference on Information and Knowledge Management}} (Glasgow,
  Scotland, UK) \emph{(\bibinfo{series}{CIKM '11})}.
  \bibinfo{publisher}{Association for Computing Machinery},
  \bibinfo{address}{New York, NY, USA}, \bibinfo{pages}{249–258}.
\newblock
\showISBNx{9781450307178}
\urldef\tempurl%
\url{https://doi.org/10.1145/2063576.2063618}
\showDOI{\tempurl}


\bibitem[\protect\citeauthoryear{Holland}{Holland}{1986}]%
        {Holland86}
\bibfield{author}{\bibinfo{person}{Paul~W Holland}.}
  \bibinfo{year}{1986}\natexlab{}.
\newblock \showarticletitle{Statistics and causal inference}.
\newblock \bibinfo{journal}{\emph{Journal of the American statistical
  Association}} \bibinfo{volume}{81}, \bibinfo{number}{396}
  (\bibinfo{year}{1986}), \bibinfo{pages}{945--960}.
\newblock


\bibitem[\protect\citeauthoryear{Iizuka, Yoneda, and Seki}{Iizuka
  et~al\mbox{.}}{2019}]%
        {Iizuka19}
\bibfield{author}{\bibinfo{person}{Kojiro Iizuka}, \bibinfo{person}{Takeshi
  Yoneda}, {and} \bibinfo{person}{Yoshifumi Seki}.}
  \bibinfo{year}{2019}\natexlab{}.
\newblock \showarticletitle{Greedy Optimized Multileaving for Personalization}.
  In \bibinfo{booktitle}{\emph{Proceedings of the 13th ACM Conference on
  Recommender Systems}} (Copenhagen, Denmark) \emph{(\bibinfo{series}{RecSys
  '19})}. \bibinfo{publisher}{Association for Computing Machinery},
  \bibinfo{address}{New York, NY, USA}, \bibinfo{pages}{413–417}.
\newblock
\showISBNx{9781450362436}
\urldef\tempurl%
\url{https://doi.org/10.1145/3298689.3347008}
\showDOI{\tempurl}


\bibitem[\protect\citeauthoryear{Imbens and Rubin}{Imbens and Rubin}{2015}]%
        {Imbens15}
\bibfield{author}{\bibinfo{person}{Guido~W. Imbens} {and}
  \bibinfo{person}{Donald~B. Rubin}.} \bibinfo{year}{2015}\natexlab{}.
\newblock \bibinfo{booktitle}{\emph{Causal Inference for Statistics, Social,
  and Biomedical Sciences: An Introduction}}.
\newblock \bibinfo{publisher}{Cambridge University Press},
  \bibinfo{address}{New York, NY, USA}.
\newblock
\showISBNx{0521885884}


\bibitem[\protect\citeauthoryear{Jannach and Jugovac}{Jannach and
  Jugovac}{2019}]%
        {Jannach19}
\bibfield{author}{\bibinfo{person}{Dietmar Jannach} {and}
  \bibinfo{person}{Michael Jugovac}.} \bibinfo{year}{2019}\natexlab{}.
\newblock \showarticletitle{Measuring the Business Value of Recommender
  Systems}.
\newblock \bibinfo{journal}{\emph{ACM Trans. Manage. Inf. Syst.}}
  \bibinfo{volume}{10}, \bibinfo{number}{4}, Article \bibinfo{articleno}{16}
  (\bibinfo{date}{Dec.} \bibinfo{year}{2019}), \bibinfo{numpages}{23}~pages.
\newblock
\showISSN{2158-656X}
\urldef\tempurl%
\url{https://doi.org/10.1145/3370082}
\showDOI{\tempurl}


\bibitem[\protect\citeauthoryear{Joachims}{Joachims}{2002}]%
        {Joachims02}
\bibfield{author}{\bibinfo{person}{Thorsten Joachims}.}
  \bibinfo{year}{2002}\natexlab{}.
\newblock \showarticletitle{Optimizing Search Engines Using Clickthrough Data}.
  In \bibinfo{booktitle}{\emph{Proceedings of the Eighth ACM SIGKDD
  International Conference on Knowledge Discovery and Data Mining}} (Edmonton,
  Alberta, Canada) \emph{(\bibinfo{series}{KDD '02})}.
  \bibinfo{publisher}{Association for Computing Machinery},
  \bibinfo{address}{New York, NY, USA}, \bibinfo{pages}{133–142}.
\newblock
\showISBNx{158113567X}
\urldef\tempurl%
\url{https://doi.org/10.1145/775047.775067}
\showDOI{\tempurl}


\bibitem[\protect\citeauthoryear{Joachims}{Joachims}{2003}]%
        {Joachims03}
\bibfield{author}{\bibinfo{person}{Thorsten Joachims}.}
  \bibinfo{year}{2003}\natexlab{}.
\newblock \showarticletitle{Evaluating Retrieval Performance Using Clickthrough
  Data}.
\newblock In \bibinfo{booktitle}{\emph{Text Mining, Theoretical Aspects and
  Applications}}, \bibfield{editor}{\bibinfo{person}{J{\"{u}}rgen Franke},
  \bibinfo{person}{Gholamreza Nakhaeizadeh}, {and} \bibinfo{person}{Ingrid
  Renz}} (Eds.). \bibinfo{publisher}{Physica-Verlag}, \bibinfo{pages}{79--96}.
\newblock


\bibitem[\protect\citeauthoryear{Lunceford and Davidian}{Lunceford and
  Davidian}{2004}]%
        {Lunceford04}
\bibfield{author}{\bibinfo{person}{Jared~K Lunceford} {and}
  \bibinfo{person}{Marie Davidian}.} \bibinfo{year}{2004}\natexlab{}.
\newblock \showarticletitle{Stratification and weighting via the propensity
  score in estimation of causal treatment effects: a comparative study}.
\newblock \bibinfo{journal}{\emph{Statistics in medicine}}
  \bibinfo{volume}{23}, \bibinfo{number}{19} (\bibinfo{year}{2004}),
  \bibinfo{pages}{2937--2960}.
\newblock


\bibitem[\protect\citeauthoryear{Ning, Desrosiers, and Karypis}{Ning
  et~al\mbox{.}}{2015}]%
        {Ning15}
\bibfield{author}{\bibinfo{person}{Xia Ning}, \bibinfo{person}{Christian
  Desrosiers}, {and} \bibinfo{person}{George Karypis}.}
  \bibinfo{year}{2015}\natexlab{}.
\newblock \bibinfo{booktitle}{\emph{A Comprehensive Survey of
  Neighborhood-Based Recommendation Methods}}.
\newblock \bibinfo{publisher}{Springer US}, \bibinfo{address}{Boston, MA},
  \bibinfo{pages}{37--76}.
\newblock


\bibitem[\protect\citeauthoryear{Radlinski and Craswell}{Radlinski and
  Craswell}{2013}]%
        {Radlinski13}
\bibfield{author}{\bibinfo{person}{Filip Radlinski} {and} \bibinfo{person}{Nick
  Craswell}.} \bibinfo{year}{2013}\natexlab{}.
\newblock \showarticletitle{Optimized Interleaving for Online Retrieval
  Evaluation}. In \bibinfo{booktitle}{\emph{Proceedings of the Sixth ACM
  International Conference on Web Search and Data Mining}} (Rome, Italy)
  \emph{(\bibinfo{series}{WSDM '13})}. \bibinfo{publisher}{Association for
  Computing Machinery}, \bibinfo{address}{New York, NY, USA},
  \bibinfo{pages}{245–254}.
\newblock
\showISBNx{9781450318693}
\urldef\tempurl%
\url{https://doi.org/10.1145/2433396.2433429}
\showDOI{\tempurl}


\bibitem[\protect\citeauthoryear{Radlinski, Kurup, and Joachims}{Radlinski
  et~al\mbox{.}}{2008}]%
        {Radlinski08}
\bibfield{author}{\bibinfo{person}{Filip Radlinski}, \bibinfo{person}{Madhu
  Kurup}, {and} \bibinfo{person}{Thorsten Joachims}.}
  \bibinfo{year}{2008}\natexlab{}.
\newblock \showarticletitle{How Does Clickthrough Data Reflect Retrieval
  Quality?}. In \bibinfo{booktitle}{\emph{Proceedings of the 17th ACM
  Conference on Information and Knowledge Management}} (Napa Valley,
  California, USA) \emph{(\bibinfo{series}{CIKM '08})}.
  \bibinfo{publisher}{Association for Computing Machinery},
  \bibinfo{address}{New York, NY, USA}, \bibinfo{pages}{43–52}.
\newblock
\showISBNx{9781595939913}
\urldef\tempurl%
\url{https://doi.org/10.1145/1458082.1458092}
\showDOI{\tempurl}


\bibitem[\protect\citeauthoryear{Rendle, Freudenthaler, Gantner, and
  Schmidt-Thieme}{Rendle et~al\mbox{.}}{2009}]%
        {Rendle09}
\bibfield{author}{\bibinfo{person}{Steffen Rendle}, \bibinfo{person}{Christoph
  Freudenthaler}, \bibinfo{person}{Zeno Gantner}, {and} \bibinfo{person}{Lars
  Schmidt-Thieme}.} \bibinfo{year}{2009}\natexlab{}.
\newblock \showarticletitle{BPR: Bayesian Personalized Ranking from Implicit
  Feedback}. In \bibinfo{booktitle}{\emph{Proceedings of the Twenty-Fifth
  Conference on Uncertainty in Artificial Intelligence}} (Montreal, Quebec,
  Canada) \emph{(\bibinfo{series}{UAI '09})}. \bibinfo{publisher}{AUAI Press},
  \bibinfo{address}{Arlington, Virginia, USA}, \bibinfo{pages}{452–461}.
\newblock
\showISBNx{9780974903958}


\bibitem[\protect\citeauthoryear{Rosenbaum and Rubin}{Rosenbaum and
  Rubin}{1983}]%
        {Rosenbaum83}
\bibfield{author}{\bibinfo{person}{Paul~R Rosenbaum} {and}
  \bibinfo{person}{Donald~B Rubin}.} \bibinfo{year}{1983}\natexlab{}.
\newblock \showarticletitle{The central role of the propensity score in
  observational studies for causal effects}.
\newblock \bibinfo{journal}{\emph{Biometrika}} \bibinfo{volume}{70},
  \bibinfo{number}{1} (\bibinfo{year}{1983}), \bibinfo{pages}{41--55}.
\newblock


\bibitem[\protect\citeauthoryear{Rubin}{Rubin}{1974}]%
        {Rubin74}
\bibfield{author}{\bibinfo{person}{Donald~B Rubin}.}
  \bibinfo{year}{1974}\natexlab{}.
\newblock \showarticletitle{Estimating causal effects of treatments in
  randomized and nonrandomized studies.}
\newblock \bibinfo{journal}{\emph{Journal of educational Psychology}}
  \bibinfo{volume}{66}, \bibinfo{number}{5} (\bibinfo{year}{1974}),
  \bibinfo{pages}{688}.
\newblock


\bibitem[\protect\citeauthoryear{Saito, Yaginuma, Nishino, Sakata, and
  Nakata}{Saito et~al\mbox{.}}{2020}]%
        {Saito20a}
\bibfield{author}{\bibinfo{person}{Yuta Saito}, \bibinfo{person}{Suguru
  Yaginuma}, \bibinfo{person}{Yuta Nishino}, \bibinfo{person}{Hayato Sakata},
  {and} \bibinfo{person}{Kazuhide Nakata}.} \bibinfo{year}{2020}\natexlab{}.
\newblock \showarticletitle{Unbiased Recommender Learning from
  Missing-Not-At-Random Implicit Feedback}. In
  \bibinfo{booktitle}{\emph{Proceedings of the 13th International Conference on
  Web Search and Data Mining}} (Houston, TX, USA) \emph{(\bibinfo{series}{WSDM
  ’20})}. \bibinfo{publisher}{Association for Computing Machinery},
  \bibinfo{address}{New York, NY, USA}, \bibinfo{pages}{501–509}.
\newblock
\showISBNx{9781450368223}
\urldef\tempurl%
\url{https://doi.org/10.1145/3336191.3371783}
\showDOI{\tempurl}


\bibitem[\protect\citeauthoryear{Sato, Izumo, and Sonoda}{Sato
  et~al\mbox{.}}{2016}]%
        {Sato16}
\bibfield{author}{\bibinfo{person}{Masahiro Sato}, \bibinfo{person}{Hidetaka
  Izumo}, {and} \bibinfo{person}{Takashi Sonoda}.}
  \bibinfo{year}{2016}\natexlab{}.
\newblock \showarticletitle{Modeling Individual Users' Responsiveness to
  Maximize Recommendation Impact}. In \bibinfo{booktitle}{\emph{Proceedings of
  the 2016 Conference on User Modeling Adaptation and Personalization}}
  (Halifax, Nova Scotia, Canada) \emph{(\bibinfo{series}{UMAP '16})}.
  \bibinfo{publisher}{ACM}, \bibinfo{address}{New York, NY, USA},
  \bibinfo{pages}{259--267}.
\newblock
\showISBNx{978-1-4503-4368-8}
\urldef\tempurl%
\url{https://doi.org/10.1145/2930238.2930259}
\showDOI{\tempurl}


\bibitem[\protect\citeauthoryear{Sato, Singh, Takemori, Sonoda, Zhang, and
  Ohkuma}{Sato et~al\mbox{.}}{2019}]%
        {Sato19}
\bibfield{author}{\bibinfo{person}{Masahiro Sato}, \bibinfo{person}{Janmajay
  Singh}, \bibinfo{person}{Sho Takemori}, \bibinfo{person}{Takashi Sonoda},
  \bibinfo{person}{Qian Zhang}, {and} \bibinfo{person}{Tomoko Ohkuma}.}
  \bibinfo{year}{2019}\natexlab{}.
\newblock \showarticletitle{Uplift-Based Evaluation and Optimization of
  Recommenders}. In \bibinfo{booktitle}{\emph{Proceedings of the 13th ACM
  Conference on Recommender Systems}} (Copenhagen, Denmark)
  \emph{(\bibinfo{series}{RecSys ’19})}. \bibinfo{publisher}{Association for
  Computing Machinery}, \bibinfo{address}{New York, NY, USA},
  \bibinfo{pages}{296–304}.
\newblock
\showISBNx{9781450362436}
\urldef\tempurl%
\url{https://doi.org/10.1145/3298689.3347018}
\showDOI{\tempurl}


\bibitem[\protect\citeauthoryear{Sato, Singh, Takemori, Sonoda, Zhang, and
  Ohkuma}{Sato et~al\mbox{.}}{2020a}]%
        {Sato20b}
\bibfield{author}{\bibinfo{person}{Masahiro Sato}, \bibinfo{person}{Janmajay
  Singh}, \bibinfo{person}{Sho Takemori}, \bibinfo{person}{Takashi Sonoda},
  \bibinfo{person}{Qian Zhang}, {and} \bibinfo{person}{Tomoko Ohkuma}.}
  \bibinfo{year}{2020}\natexlab{a}.
\newblock \showarticletitle{Modeling User Exposure with Recommendation
  Influence}. In \bibinfo{booktitle}{\emph{Proceedings of the 35th Annual ACM
  Symposium on Applied Computing}} (Brno, Czech Republic)
  \emph{(\bibinfo{series}{SAC '20})}. \bibinfo{publisher}{Association for
  Computing Machinery}, \bibinfo{address}{New York, NY, USA},
  \bibinfo{pages}{1461–1464}.
\newblock
\showISBNx{9781450368667}
\urldef\tempurl%
\url{https://doi.org/10.1145/3341105.3375760}
\showDOI{\tempurl}


\bibitem[\protect\citeauthoryear{Sato, Takemori, Singh, and Ohkuma}{Sato
  et~al\mbox{.}}{2020b}]%
        {Sato20}
\bibfield{author}{\bibinfo{person}{Masahiro Sato}, \bibinfo{person}{Sho
  Takemori}, \bibinfo{person}{Janmajay Singh}, {and} \bibinfo{person}{Tomoko
  Ohkuma}.} \bibinfo{year}{2020}\natexlab{b}.
\newblock \showarticletitle{Unbiased Learning for the Causal Effect of
  Recommendation}. In \bibinfo{booktitle}{\emph{Fourteenth ACM Conference on
  Recommender Systems}} (Virtual Event, Brazil) \emph{(\bibinfo{series}{RecSys
  '20})}. \bibinfo{publisher}{Association for Computing Machinery},
  \bibinfo{address}{New York, NY, USA}, \bibinfo{pages}{378–387}.
\newblock
\showISBNx{9781450375832}
\urldef\tempurl%
\url{https://doi.org/10.1145/3383313.3412261}
\showDOI{\tempurl}


\bibitem[\protect\citeauthoryear{Sato, Takemori, Singh, and Zhang}{Sato
  et~al\mbox{.}}{2021}]%
        {Sato21}
\bibfield{author}{\bibinfo{person}{Masahiro Sato}, \bibinfo{person}{Sho
  Takemori}, \bibinfo{person}{Janmajay Singh}, {and} \bibinfo{person}{Qian
  Zhang}.} \bibinfo{year}{2021}\natexlab{}.
\newblock \showarticletitle{Causality-Aware Neighborhood Methods for
  Recommender Systems}.
\newblock  (\bibinfo{year}{2021}), \bibinfo{pages}{603--618}.
\newblock
\showISBNx{978-3-030-72113-8}
\urldef\tempurl%
\url{https://doi.org/10.1007/978-3-030-72113-8_40}
\showDOI{\tempurl}


\bibitem[\protect\citeauthoryear{Schnabel, Swaminathan, Singh, Chandak, and
  Joachims}{Schnabel et~al\mbox{.}}{2016}]%
        {Schnabel16}
\bibfield{author}{\bibinfo{person}{Tobias Schnabel}, \bibinfo{person}{Adith
  Swaminathan}, \bibinfo{person}{Ashudeep Singh}, \bibinfo{person}{Navin
  Chandak}, {and} \bibinfo{person}{Thorsten Joachims}.}
  \bibinfo{year}{2016}\natexlab{}.
\newblock \showarticletitle{Recommendations as Treatments: Debiasing Learning
  and Evaluation}. In \bibinfo{booktitle}{\emph{Proceedings of the 33rd
  International Conference on International Conference on Machine Learning -
  Volume 48}} (New York, NY, USA) \emph{(\bibinfo{series}{ICML’16})}.
  \bibinfo{publisher}{JMLR.org}, \bibinfo{pages}{1670–1679}.
\newblock


\bibitem[\protect\citeauthoryear{Schuth, Bruintjes, Bu\"{u}ttner, van Doorn,
  Groenland, Oosterhuis, Tran, Veeling, van~der Velde, Wechsler, Woudenberg,
  and de~Rijke}{Schuth et~al\mbox{.}}{2015}]%
        {Schuth15}
\bibfield{author}{\bibinfo{person}{Anne Schuth}, \bibinfo{person}{Robert-Jan
  Bruintjes}, \bibinfo{person}{Fritjof Bu\"{u}ttner}, \bibinfo{person}{Joost
  van Doorn}, \bibinfo{person}{Carla Groenland}, \bibinfo{person}{Harrie
  Oosterhuis}, \bibinfo{person}{Cong-Nguyen Tran}, \bibinfo{person}{Bas
  Veeling}, \bibinfo{person}{Jos van~der Velde}, \bibinfo{person}{Roger
  Wechsler}, \bibinfo{person}{David Woudenberg}, {and} \bibinfo{person}{Maarten
  de Rijke}.} \bibinfo{year}{2015}\natexlab{}.
\newblock \showarticletitle{Probabilistic Multileave for Online Retrieval
  Evaluation}. In \bibinfo{booktitle}{\emph{Proceedings of the 38th
  International ACM SIGIR Conference on Research and Development in Information
  Retrieval}} (Santiago, Chile) \emph{(\bibinfo{series}{SIGIR '15})}.
  \bibinfo{publisher}{Association for Computing Machinery},
  \bibinfo{address}{New York, NY, USA}, \bibinfo{pages}{955–958}.
\newblock
\showISBNx{9781450336215}
\urldef\tempurl%
\url{https://doi.org/10.1145/2766462.2767838}
\showDOI{\tempurl}


\bibitem[\protect\citeauthoryear{Schuth, Sietsma, Whiteson, Lefortier, and
  de~Rijke}{Schuth et~al\mbox{.}}{2014}]%
        {Schuth14}
\bibfield{author}{\bibinfo{person}{Anne Schuth}, \bibinfo{person}{Floor
  Sietsma}, \bibinfo{person}{Shimon Whiteson}, \bibinfo{person}{Damien
  Lefortier}, {and} \bibinfo{person}{Maarten de Rijke}.}
  \bibinfo{year}{2014}\natexlab{}.
\newblock \showarticletitle{Multileaved Comparisons for Fast Online
  Evaluation}. In \bibinfo{booktitle}{\emph{Proceedings of the 23rd ACM
  International Conference on Information and Knowledge Management}} (Shanghai,
  China) \emph{(\bibinfo{series}{CIKM '14})}. \bibinfo{publisher}{Association
  for Computing Machinery}, \bibinfo{address}{New York, NY, USA},
  \bibinfo{pages}{71–80}.
\newblock
\showISBNx{9781450325981}
\urldef\tempurl%
\url{https://doi.org/10.1145/2661829.2661952}
\showDOI{\tempurl}


\bibitem[\protect\citeauthoryear{Sharma, Hofman, and Watts}{Sharma
  et~al\mbox{.}}{2015}]%
        {Sharma15}
\bibfield{author}{\bibinfo{person}{Amit Sharma}, \bibinfo{person}{Jake~M.
  Hofman}, {and} \bibinfo{person}{Duncan~J. Watts}.}
  \bibinfo{year}{2015}\natexlab{}.
\newblock \showarticletitle{Estimating the Causal Impact of Recommendation
  Systems from Observational Data}. In \bibinfo{booktitle}{\emph{Proceedings of
  the Sixteenth ACM Conference on Economics and Computation}} (Portland,
  Oregon, USA) \emph{(\bibinfo{series}{EC '15})}. \bibinfo{publisher}{ACM},
  \bibinfo{address}{New York, NY, USA}, \bibinfo{pages}{453--470}.
\newblock
\showISBNx{978-1-4503-3410-5}
\urldef\tempurl%
\url{https://doi.org/10.1145/2764468.2764488}
\showDOI{\tempurl}


\bibitem[\protect\citeauthoryear{Wang, Liang, Charlin, and Blei}{Wang
  et~al\mbox{.}}{2020}]%
        {Wang20}
\bibfield{author}{\bibinfo{person}{Yixin Wang}, \bibinfo{person}{Dawen Liang},
  \bibinfo{person}{Laurent Charlin}, {and} \bibinfo{person}{David~M. Blei}.}
  \bibinfo{year}{2020}\natexlab{}.
\newblock \showarticletitle{Causal Inference for Recommender Systems}. In
  \bibinfo{booktitle}{\emph{Fourteenth ACM Conference on Recommender Systems}}
  (Virtual Event, Brazil) \emph{(\bibinfo{series}{RecSys '20})}.
  \bibinfo{publisher}{Association for Computing Machinery},
  \bibinfo{address}{New York, NY, USA}, \bibinfo{pages}{426–431}.
\newblock
\showISBNx{9781450375832}
\urldef\tempurl%
\url{https://doi.org/10.1145/3383313.3412225}
\showDOI{\tempurl}


\end{thebibliography}

\end{document}